\documentclass[runningheads]{llncs}
\usepackage[T1]{fontenc}
\usepackage{booktabs}
\usepackage{multirow}
\usepackage{hyperref}
\usepackage{amsmath,amssymb,amsfonts}
\usepackage{algorithmic}
\usepackage{graphicx}
\usepackage{textcomp}
\usepackage{xcolor}
\usepackage{graphicx}
\begin{document}
\title{Mamba Based Feature Extraction And Adaptive Multilevel Feature Fusion For 3D Tumor Segmentation From Multi-modal Medical Image}
%
%
\author{Zexin Ji\inst{1,2,4}\and
Beiji Zou\inst{1,2}\and
Xiaoyan Kui\inst{1,2}\and
Hua Li\inst{3}\and
Pierre Vera\inst{4}\and
Su Ruan\inst{5}
}
\authorrunning{F. Author et al.}
%
\institute{School of Computer Science and Engineering, Central South University, Changsha, 410083, China \and
Hunan Engineering Research Center of Machine Vision and Intelligent Medicine, Central South University, Changsha, 410083, China 
\and
Department of Radiation Oncology, Washington University in St. Louis, USA
\and
Department of Nuclear Medicine, Henri Becquerel Cancer Center, Rouen, France\and
University of Rouen-Normandy, AMIS - QuantIF UR 4108, F-76000, Rouen, France
\\
\email{\{zexin.ji\}@csu.edu.cn}}

\maketitle              
\begin{abstract}
Multi-modal 3D medical image segmentation aims to accurately identify tumor regions across different modalities, facing challenges from variations in image intensity and tumor morphology. Traditional convolutional neural network (CNN)-based methods struggle with capturing global features, while Transformers-based methods, despite effectively capturing global context, encounter high computational costs in 3D medical image segmentation. The Mamba model combines linear scalability with long-distance modeling, making it a promising approach for visual representation learning. However, Mamba-based 3D multi-modal segmentation still struggles to leverage modality-specific features and fuse complementary information effectively. In this paper, we propose a Mamba based feature extraction and adaptive multilevel feature fusion for 3D tumor segmentation using multi-modal medical image. We first develop the specific modality Mamba encoder to efficiently extract long-range relevant features that represent anatomical and pathological structures present in each modality. Moreover, we design an bi-level synergistic integration block that dynamically merges multi-modal and multi-level complementary features by the modality attention and channel attention learning. Lastly, the decoder combines deep semantic information with fine-grained details to generate the tumor segmentation map. Experimental results on medical image datasets (PET/CT and MRI multi-sequence) show that our approach achieve competitive performance compared to the state-of-the-art CNN, Transformer, and Mamba-based approaches.
\keywords{3D medical image  \and Tumor segmentation \and Multi-modal feature fusion \and Mamba.}
\end{abstract}
\section{Introduction}
Medical image segmentation is one of the most important tasks in the field of medical image analysis. By segmenting organs, tissues, or lesions in medical images, physicians can more clearly identify affected areas, thereby improving the accuracy of disease diagnosis and therapy. In clinical practice, physicians often use different modalities to better understand patient conditions. For example, T1-weighted (T1) shows brain anatomy, while T1 with gadolinium contrast (T1-Gd) highlights enhanced lesions with a contrast agent. T2-weighted (T2) detects edema, and FLAIR suppresses cerebrospinal fluid signals to emphasize abnormalities like lesions. Leveraging advancements in deep learning, multi-modal 3D medical image segmentation tasks have experienced significant progress in recent years, and numerous CNN-based methods~\cite{myronenko20193d,zhou2020multi,lee20223d,roy2023mednext,huang2023application} have been proposed to enhance the segmentation performance. 
For example, an automated brain tumor segmentation network~\cite{myronenko20193d} based on an encoder-decoder architecture with a variational auto-encoder branch for regularization won 1st place in the BraTS 2018 challenge. 
However, the CNN-based approaches struggle to capture long-range dependencies due to their local receptive fields. The Vision Transformer model (ViT)~\cite{DBLP:conf/iclr/DosovitskiyB0WZ21} has gained significant interest for its ability to capture long-range dependencies through the self-attention mechanism.
This has led to many remarkable advancements in the multi-modal medical image segmentation task~\cite{hatamizadeh2022unetr,hatamizadeh2021swin,he2023swinunetr,xiao2023transformers}.
Hatamizadeh \emph{et al.}~\cite{hatamizadeh2021swin} developed Swin UNETR, a 3D brain tumor segmentation model that utilizes a Swin transformer as the encoder.
However, Transformers significant computational resources, and their quadratic complexity in sequence length makes them inefficient for long inputs.
Recently, state space models (SSMs)~\cite{gu2021combining}, in particular Mamba~\cite{gu2023mamba}, has shown superior performance in long-range modeling compared to Transformers and demonstrates linear scalability with sequence length. Vision Mamba represents an adaptation of the Mamba architecture, specifically designed to better accommodate complex computer vision tasks, including image super-resolution~\cite{ji2024deform,ji2025generation}, reconstruction~\cite{huang2025enhancing,wang2025cdrmamba}, classification~\cite{xing2024segmamba,li2024mambahsi}, and segmentation~\cite{xing2024segmamba}. For example, Xing \emph{et al.}~\cite{xing2024segmamba} proposed the SegMamba model, which integrates Mamba model to achieve 3D medical image segmentation.

Although the above methods also adopt multi-modal MRI data to improve brain tumor segmentation performance, they still suffer from the following limitations:  
(1) Most methods treat multi-modal data as stacked input channels and rely on the network to implicitly learn relationships among modalities, without explicitly modeling modality-specific characteristics and their interactions.
(2) While each modality provides unique and complementary information, current approaches lack dedicated mechanisms to highlight modality-specific strengths and effectively integrate complementary features.
(3) Many methods rely on fixed or early fusion strategies, which lack the flexibility to dynamically balance the contribution of each modality throughout the feature extraction process.
\textit{Therefore, a challenge that also remains for multi-modal medical image segmentation tasks is how to simultaneously leverage modality-specific features and effectively fuse complementary features to achieve efficient segmentation.}

\begin{figure*}[t]
    \centering
    \includegraphics[width=0.9\linewidth]{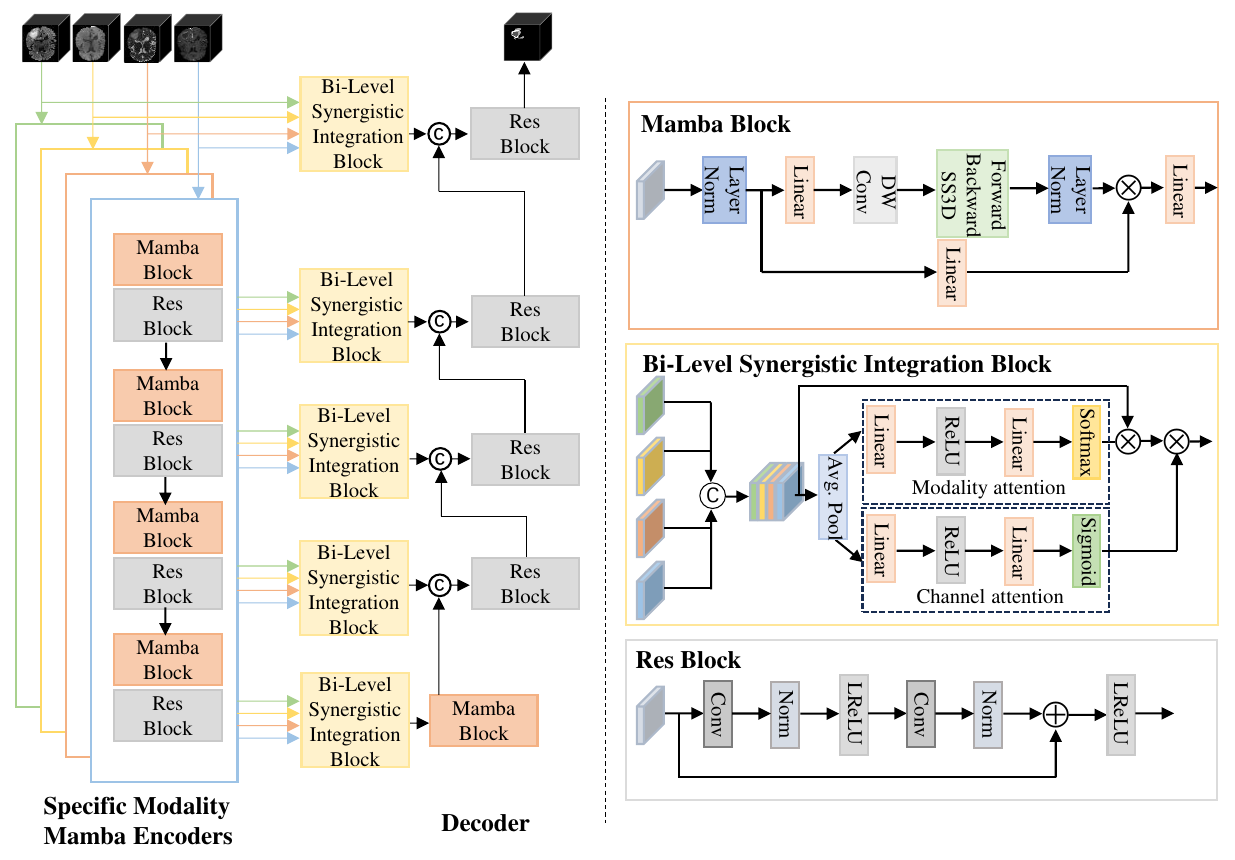}
    \caption{Our architecture mainly includes Mamba block, bi-level synergistic integration block, and Res block, shown through a brain tumor segmentation with 4-sequence MRI.}
    \label{fig:Framework}
\end{figure*}

In this paper, we propose a Mamba based feature extraction and adaptive multilevel feature fusion for 3D tumor segmentation from multi-modal medical image. 
Enhancing tumor features in a specific modality and fusing complementary features across modalities helps capture differences in tumor appearance and location among patients.
To achieve this, we design the network which specifically consist of the specific modality Mamba encoder, bi-level synergistic integration block, and the decoder.
Specifically, different 3D modalities are input into each specific modality Mamba encoder.
The Mamba encoder independently extracts long-range relevant features, capturing spatial and contextual information to preserve and emphasize unique characteristics of each modality before fusion.
Unlike simple concatenation for fusing multi-modal information, we design the bi-level synergistic integration block to dynamically explore the complementary information between modalities. Finally, the decoder combines high-level semantic information from deep layers with fine-grained details from shallower layers to produce the segmentation map delineating the tumor regions. 
Fig.\ref{fig:Framework} illustrates the pipeline of our model. Quantitative and qualitative experimental results demonstrate that our approach has competitive performance compared to state-of-the-art approaches.

The main contributions of our approach are summarized as follows:

\begin{itemize}
    \item We propose a Mamba-based approach for feature extraction and adaptive multilevel feature fusion in 3D tumor segmentation using multi-modal medical image.
   
    \item We propose the specific modality Mamba encoder to efficiently extract contextual characteristics of different modality. The bi-level synergistic integration block is designed to dynamically capture the complementary information across modalities.

    \item We conduct extensive experiments on BraTS2023 (MRI) for brain tumor segmentation and Hecktor2022 (PET/CT) for head and neck tumor segmentation, demonstrating the effectiveness and robustness of the proposed method compared to existing approaches.

\end{itemize}

The structure of the paper is as follows: Section \ref{sec:Related} reviews the state-of-the-art segmentation methods. Section \ref{sec:METHOD} focuses on detailing the proposed approach. In Section \ref{sec:EXPERIMENTS}, we outline the experimental components. Finally, the conclusion is provided in Section \ref{sec:CONCLUSIONS}.

\section{METHOD}
\label{sec:METHOD}

\subsection{Specific Modality Mamba Encoder}

Tumors exhibit distinct appearances across modalities. For instance, T1-Gd highlights enhancing tumor regions, while FLAIR is sensitive to edema. Moreover, different modalities have varying noise levels and contrasts. Most existing methods simply stack multi-modal data as input channels and rely on the network to implicitly capture relationships, without explicitly modeling the unique characteristics of each modality.
Therefore, we use specific modality Mamba encoder to realize specialized feature extraction for each modality, capturing the most relevant tumor-related information and reducing the impact of modality-specific artifacts or inconsistencies.

The Mamba characterizes state representations and predicts future states. The transformation from a 1-D function or sequence $x(t) \in \mathbb{R}$ to an output $y(t) \in \mathbb{R}$ is achieved through a hidden state $h(t) \in \mathbb{R}^{\mathbb{N}}$.
The implementation is typically achieved through linear ordinary differential equations (ODEs)~\cite{osborne1968note}, as follows.

\begin{equation}\label{eq01}
    h^{\prime}(t)=\mathbf{A} h(t)+\mathbf{B} x(t), y(t)=\mathbf{C} h(t),
\end{equation}
where $\mathbf{A} \in \mathbb{R}^{\mathrm{N} \times \mathrm{N}}$ represents the state matrix. $\mathbf{B} \in \mathbb{R}^{\mathrm{N} \times 1}$ and $\mathbf{C} \in \mathbb{R}^{1 \times \mathrm{N}}$ denote the projection parameters. The zero-order hold (ZOH)~\cite{galias2008analysis} is applied to discretize ODEs, making it more suitable for deep learning. 
Using a timescale parameter $\Delta$, it transforms continuous-time matrices $\mathbf{A}$ and $\mathbf{B}$ into discrete forms, $\overline{\mathbf{A}}$ and $\overline{\mathbf{B}}$. This discretization is carried out with the following steps:

\begin{equation}\label{eq2}
\overline{\mathbf{A}}=\exp (\Delta \mathbf{A}), \overline{\mathbf{B}}=(\Delta \mathbf{A})^{-1}(\exp (\Delta \mathbf{A})-\mathbf{I}) \cdot \Delta \mathbf{B} .
\end{equation}

After discretization, Eq.\ref{eq01} takes a form suitable for discrete-time processing:

\begin{equation}\label{eq3}
h_t=\overline{\mathbf{A}} h_{t-1}+\overline{\mathbf{B}} x_t, y_t=\mathbf{C} h_t .
\end{equation}

Mamba introduces selective structured state space sequence models (S6), allowing dynamic parameterization, where parameters $\overline{\mathbf{B}}$, $\mathbf{C}$, and $\Delta$ are determined by input data for unique adaptation. The Mamba block in Fig.\ref{fig:Framework} processes image patches by splitting the input into two paths after layer normalization. One path uses a linear layer, while the other applies a linear layer, a depthwise separable convolution (DW Conv), 3D-Selective-Scan (SS3D), and layer normalization. The two paths are merged via multiplication and a linear layer.
The SS3D is designed to effectively utilize input features from forward and backward perspectives, enhancing the capacity for the extraction of contextual information. 
Then we also incorporate the Res block after the Mamba block, which is specifically employed to further extract fine-grained spatial details and refine feature representation.
This combination allows the network to capture both global context and intricate local features.

\subsection{Bi-Level Synergistic Integration Block}

Due to the reliance on fixed or early fusion strategies, many methods lack the flexibility to dynamically balance the contributions of each modality throughout the feature extraction process.
To address this limitation, the Fig.\ref{fig:Framework} shows the proposed bi-level synergistic integration block.
This block can flexibly integrate features from different modalities using a combination of global modality weights and channel-specific weights. This design allows the model to assign varying importance to each modality and its feature channels, ensuring an adaptive and balanced fusion of multi-modal information.

Given the input features from $M$ modalities, we denote them as:
\[
\mathbf{X} = \left\{ \mathbf{X}^{(1)}, \mathbf{X}^{(2)}, \dots, \mathbf{X}^{(M)} \right\}
\]
where $\mathbf{X}^{(m)} \in \mathbb{R}^{C \times H \times W \times D}$ represents the feature map of the $m$-th modality.

First, all modality-specific features are concatenated along the modality dimension:
\[
\mathbf{X}_{\text{concat}} = \text{Concat} \left( \mathbf{X}^{(1)}, \mathbf{X}^{(2)}, \dots, \mathbf{X}^{(M)} \right)
\]
Then, an average pooling operation is applied to obtain a compact global descriptor:
\[
\mathbf{X}_{\text{pool}} = \text{AvgPool} \left( \mathbf{X}_{\text{concat}} \right)
\]

Next, the bi-level synergistic integration block computes modality-level attention weights to adaptively emphasize important modalities. This is achieved through a fully connected layer followed by a ReLU activation and another linear projection, finalized by a softmax function:
\[
\mathbf{A}_{\text{modality}} = \text{Softmax} \left( W_2^{\text{mod}} \cdot \sigma \left( W_1^{\text{mod}} \cdot \mathbf{X}_{\text{pool}} \right) \right)
\]
where $W_1^{\text{mod}}$ and $W_2^{\text{mod}}$ are learnable weight matrices, and $\sigma(\cdot)$ denotes the ReLU activation function.

In parallel, channel-level attention is computed to refine the feature representation within each modality. Similarly, it employs two linear transformations with ReLU activation, followed by a sigmoid function to generate channel-specific weights:
\[
\mathbf{A}_{\text{channel}} = \sigma \left( W_2^{\text{ch}} \cdot \sigma \left( W_1^{\text{ch}} \cdot \mathbf{X}_{\text{pool}} \right) \right)
\]
where $W_1^{\text{ch}}$ and $W_2^{\text{ch}}$ are the learnable parameters for channel attention.

Subsequently, the modality and channel attention weights are applied to recalibrate the input features:
\[
\mathbf{X}_{\text{out}}^{(m)} = \mathbf{A}_{\text{modality}}^{(m)} \cdot \left( \mathbf{A}_{\text{channel}} \odot \mathbf{X}^{(m)} \right)
\]
where $\odot$ represents element-wise multiplication along the channel dimension.

This bi-level attention design enables the network to effectively prioritize critical modalities and feature channels, thereby capturing rich texture details and accurately delineating tumor boundaries in multi-modal brain tumor segmentation tasks.

\subsection{Decoder}

Notably, instead of directly proceeding to the upsampling stage after the bottleneck, our approach integrates a specialized Mamba Block.
While the bottleneck compresses features to capture global representations, this additional block compensates for the loss of spatial information by selectively amplifying critical features. This selective enhancement strengthens the reliability of segmentation, especially in challenging cases like complex tumor regions.

As illustrated in Fig.\ref{fig:Framework}, the decoder is designed to progressively restore the spatial resolution of the feature maps. 
This process is achieved through a combination of upsampling layers and the integration of multi-scale information from the encoder via skip connections. These skip connections ensure that fine-grained spatial details captured in the encoder are preserved and further incorporated into the decoder.
The feature aggregation step within the decoder is performed through the Res block to effectively combine high-level semantic information with low-level spatial details, ensuring a richer representation of features for accurate tumor segmentation.

\section{EXPERIMENTS}
\label{sec:EXPERIMENTS}

\subsection{Dataset}
We conducted experiments on the {BraTS2023} dataset\footnote{https://www.med.upenn.edu/cbica/brats/} for brain tumor segmentation, which includes multiple MRI modalities: T1, T1-Gd, T2, and FLAIR. 
The dataset offers detailed annotations for three tumor categories: Whole Tumor (WT), Enhancing Tumor (ET), and Tumor Core (TC).  
To further validate the robustness of our approach across different anatomical regions, we also applied our method to the Hecktor2022 dataset\footnote{https://hecktor.grand-challenge.org/}, focusing on head and neck tumor segmentation with PET and CT.

\subsection{Evaluation Metrics}

The segmentation performance was evaluated using two common metrics: 
Dice Score and Hausdorff Distance. 

(1) The Dice Score measures the overlap between the predicted ($P$) and ground truth ($G$) tumor regions. It is defined as:

\begin{equation}
\text{Dice Score} = \frac{2 \cdot |P \cap G|}{|P| + |G|},
\end{equation}
where $|P \cap G|$ denotes the number of overlapping voxels between the prediction and the ground truth, and $|P|$ and $|G|$ represent the total number of voxels in the predicted and ground truth segmentation masks, respectively. The Dice score ranges from 0 to 1, with a higher score indicating better tumor segmentation performance.

(2) Hausdorff Distance quantifies the the maximum boundary deviation between the segmentation result and the ground truth. 
It is defined as:

\begin{equation}
HD(P, G) = \max \left\{ \max_{p \in P} \min_{g \in G} d(p, g), \max_{g \in G} \min_{p \in P} d(p, g) \right\},
\end{equation}
where $d(p, g)$ is the Euclidean distance between points $p \in P$ and $g \in G$. The smaller the Hausdorff distance indicating higher tumor segmentation accuracy.

\subsection{Implementation Details}
Our approach is implemented in PyTorch on an NVIDIA RTX A6000. We use SGD optimizer with a learning rate of 1e-3 and a weight decay of 1e-5. The network is trained with a cross-entropy loss for 1000 epochs, with a batch size of 1. We split BraTS2023 into 70\% training, 10\% validation, and 20\% testing, and Hecktor2022 into 60\% training, 20\% validation, and 20\% testing. For the BraTS2023 dataset, the evaluation metrics for comparison methods are sourced from \cite{xing2024segmamba}. 
For the Hecktor2022 dataset, we use the centrally cropped region for testing.

\begin{table*}[t]
\centering
\caption{Comparison of different network
 components.
}\label{table1}
\setlength{\tabcolsep}{6pt} 
\begin{tabular}{lcccccccc}
\toprule
\multirow{2}{*}{Methods} & \multicolumn{4}{c}{Dice Score (\%, ↑)} & \multicolumn{4}{c}{Hausdorff Distance (mm, ↓)} \\ 
                         & WT  & TC  & ET  & Mean & WT  & TC  & ET  & Mean \\ 
\midrule
SingleModality            & 90.98 & 79.12 & 62.68 & 77.59 & 3.50 & 6.45 & 8.21 & 6.05 \\
SimpleFusion             & 92.17 & 87.03 & 82.52 & 87.24 & 4.88 & 6.74 & 5.70 & 5.77 \\
MambaEncoder      & 93.17 & 90.79 & 86.45 & 90.13& 3.64 & 3.62 & 4.09 & 3.78 \\
Ours  & \textbf{94.86} & \textbf{93.68} & \textbf{87.92} & \textbf{92.15} & \textbf{2.09} & \textbf{2.49} & \textbf{3.30} & \textbf{2.62} \\

\bottomrule
\end{tabular}
\end{table*}

\subsection{Experimental results}

\subsubsection{\textbf{Ablation Study}}
To justify the components in our approach, we conducted ablation studies on the BraTS2023 dataset with the following baselines:
1) \textit{SingleModality}: A single U-Net model designed to process single-modality input;
2) \textit{SimpleFusion}: A multi-modality U-Net model where features from different modalities are fused using simple summation to perform segmentation; 3) \textit{MambaEncoder}: The Mamba is employed as the encoder to effectively capture richer global context from the input multi-modal data; 4) \textit{Ours}: Based on \textit{MambaEncoder}, \textit{Ours} adds bi-level synergistic integration block to further enhance the feature representation.
Tab.\ref{table1} shows the quantitative results, demonstrating that the proposed approach yields improved performance across all metrics.
Specifically, 
\textit{SimpleFusion} significantly outperforms \textit{SingleModality}, highlighting the importance of utilizing multimodal MRI data. Single-modality inputs (e.g., T1, T1-Gd, T2, and FLAIR) are limited in their ability to provide a comprehensive description of tumor regions, whereas multi-modal MRI offers complementary tumor information.
\textit{MambaEncoder} enables the model to capture global contextual information, addressing the limitations of traditional U-Net encoders in handling the non-local characteristics of complex tumor regions.
While \textit{MambaEncoder} excels in global feature extraction, it falls short in effectively fusing features from different MRI modalities. \textit{Ours} further adds an bi-level synergistic integration block, which dynamically learns how to integrate complementary information from various modalities based on the demands of the segmentation task. Compared to simple feature summation, this adaptive fusion mechanism prioritizes the most informative features from each modality, resulting in a Dice score improvement of 2.02 and a Hausdorff Distance reduction of 1.16.
In summary, our approach exploits the effectiveness
integrated with both the multi-modality, the specific modality Mamba encoder, and bi-level synergistic integration block.

\begin{table*}[t]
\centering
\caption{Comparison of segmentation results with state-of-the-art methods on BraTS2023. }\label{table2}
\setlength{\tabcolsep}{6pt} 
\begin{tabular}{lcccccccc}
\toprule
\multirow{2}{*}{Methods} & \multicolumn{4}{c}{Dice Score (\%, ↑)} & \multicolumn{4}{c}{Hausdorff Distance (mm, ↓)} \\ 
                         & WT  & TC  & ET  & Mean & WT  & TC  & ET  & Mean \\ 
\midrule
SegresNet             & 92.02 & 89.10 & 83.66 & 88.26 & 4.07 & 4.08 & 3.88 & 4.01 \\
UX-Net       & 93.13 & 90.03 & 85.91 & 89.69 & 4.56 & 5.68 & 4.19 & 4.81 \\
MedNeXt  & 92.41 & 87.75 & 83.96 & 88.04 & 4.98 & 4.67 & 4.51 & 4.72 \\
UNETR & 92.19 & 86.39 & 84.48 & 87.68 & 6.17 & 5.29 & 5.03 & 5.49 \\
SwinUNETR & 92.71 & 87.79 & 84.21 & 88.23 & 5.22 & 4.42 & 4.48 & 4.70 \\
SwinUNETR-V2 & 93.35 & 89.65 & 85.17 & 89.39 & 5.01 & 4.41 & 4.41 & 4.51 \\
SegMamba & 93.61 & 92.65 & 87.71 & 91.32 & 3.37 & 3.85 & 3.48 & 3.56 \\
Ours & \textbf{94.86} & \textbf{93.68} & \textbf{87.92} & \textbf{92.15} & \textbf{2.09} & \textbf{2.49} & \textbf{3.30} & \textbf{2.62} \\
\bottomrule
\end{tabular}
\end{table*}

\begin{table}[t]
\centering
\caption{Comparison of segmentation results on Hecktor2022.
}\label{table3}
\begin{tabular}{lcccccccc}
\toprule
{Methods} & {Dice Score (\%, ↑)} & {Hausdorff Distance (mm, ↓)} \\ 
                         
\midrule
PET-only            & 33.07 & 14.46 \\
Ours  & \textbf{37.12} & \textbf{12.06} \\

\bottomrule
\end{tabular}
\end{table}

\begin{figure*}[t]
    \centering
    \includegraphics[width=1\linewidth]{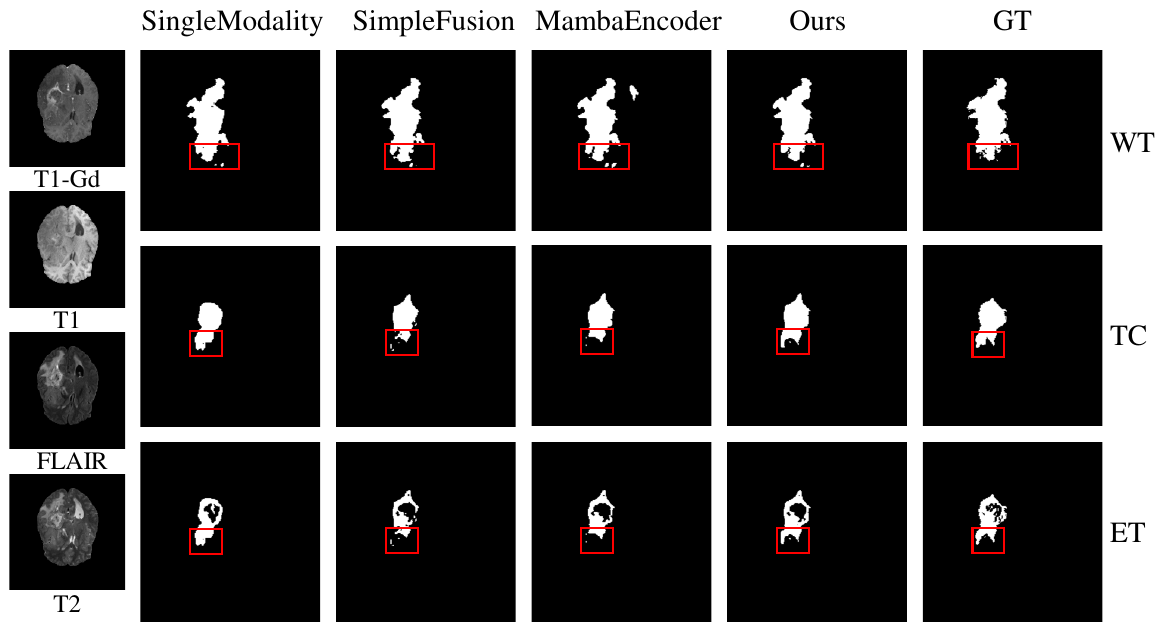}
    \caption{Qualitative results on BraTS2023 dataset.}
    \label{fig:ISBIBrats}
\end{figure*}

\begin{figure*}[t]
    \centering
    \includegraphics[width=0.95\linewidth]{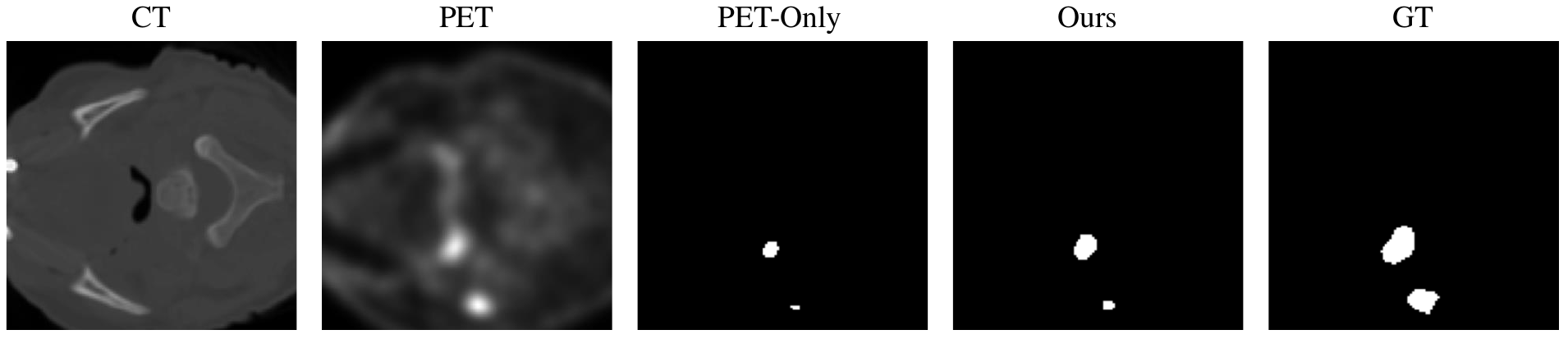}
    \caption{Qualitative results on Hecktor2022 dataset.}
    \label{fig:hecktor}
\end{figure*}

\subsubsection{\textbf{Quantitative results}}

Our approach is also compared with the state-of-the-art CNN-based (SegresNet~\cite{myronenko20193d}, UX-Net~\cite{lee20223d},  MedNeXt~\cite{roy2023mednext}), Transformer-based (UNETR~\cite{hatamizadeh2022unetr}, SwinUNETR~\cite{hatamizadeh2021swin}, SwinUNETRV2~\cite{he2023swinunetr}), and Mamba-based (SegMamba~\cite{xing2024segmamba}) approaches validated on BraTS2023 dataset. 
The quantitative experimental results on BraTS2023 dataset are summarized in Tab.~\ref{table2}.
It can be seen that our approach outperforms all segmentation approaches.
Compared to the whole tumor (WT) and tumor core (TC) regions, enhancing tumors (ET) often exhibit smaller size, heterogeneous appearance, and unclear boundaries, leading to higher segmentation errors in previous methods. 
Our approach yields a higher Dice Score and lower Hausdorff Distance in the ET region, which demonstrates robustness in capturing fine-grained details and delineating tumor boundaries,
The reasons can mainly be attributed to the following aspects: Each 3D modality is processed independently by a dedicated Mamba encoder, which captures long-range spatial and contextual features while preserving modality-specific characteristics. Furthermore, the bi-level synergistic integration block dynamically integrates complementary information across modalities. Finally, the decoder effectively combines high-level semantics with fine-grained details to generate accurate tumor segmentation maps.

To further verify the effectiveness of our method, we conducted experiments on Hecktor2022 dataset using PET/CT.
Two different segmentation approaches were compared: network using PET-only input and with multi-modality (Ours). Tab.~\ref{table3} indicates that the fusion of PET and CT outperforms the PET-only segmentation.
PET imaging excels at capturing metabolic activity, which is crucial for identifying regions with abnormal cellular function, such as tumors. However, PET images often suffer from low spatial resolution and lack detailed anatomical structures, which can make it challenging to accurately delineate tumor boundaries.
CT imaging provides high-resolution anatomical details, enabling precise localization of structures and better definition of tumor margins. By combining PET and CT, our approach benefits from both the metabolic specificity of PET and the anatomical precision of CT to improve the segmentation performance.

\subsubsection{\textbf{Qualitative results}}
The qualitative experimental results on BraTS2023 dataset can be seen in Fig.\ref{fig:ISBIBrats}. 
We can observe that our approach can clearly segment the tumor region compared to other methods. Specifically, our approach can accurately capture the fine-grained details of the enhancing tumor (ET) region while maintaining the integrity of the whole tumor (WT) and tumor core (TC) regions. Thanks to the combined use of modality-specific Mamba-based encoders and the bi-level synergistic integration block, our approach effectively preserves and leverages complementary information across modalities.
Moreover, the decoder integrates high-level semantic features with low-level spatial details to better identify various types and forms of tumors. These qualitative results further validate the quantitative findings presented earlier.

Fig.\ref{fig:hecktor} also provides qualitative comparisons using different input modalities on Hecktor2022 dataset. 
It is evident from the visual comparisons that our approach yields more complete tumor segmentation boundaries, with a better delineation of tumor regions compared to the PET-only method. 
The reason is that our approach effectively leverages metabolic specificity of PET with precise anatomical context of CT. 

\section{CONCLUSIONS}
\label{sec:CONCLUSIONS}

In this paper, we have developed a Mamba based feature extraction and adaptive multilevel feature fusion for 3D tumor segmentation from multi-modal medical image.
Our model is designed to maximize the utilization of tumor-specific information within each modality, while also leveraging complementary information across multiple modalities in an adaptive manner.
Quantitatively, our method achieves superior performance in key metrics such as Dice Score and Hausdorff Distance. 
Qualitatively, the visualizations of segmentation results from the MRI multi-sequence and PET/CT datasets confirm that our approach accurately delineates tumor boundaries, effectively capturing the unique tumor characteristics inherent to each modality.

Future research could focus on enhancing the interpretability of the model, enabling clinicians to intuitively comprehend the underlying rationale of the decision-making process of the model. 
For instance, we will mathematically explore and quantify the contribution of features from different modalities in the tumor segmentation process and their specific impact on the final outcomes.

\noindent\textbf{Acknowledgements.}
The work was supported by the National Key R$\&$D Program of China (No.2018AAA0102100); the National Natural Science Foundation of China (Nos.U22A2034, 62177047); the Key Research and Development Program of Hunan Province (No.2022SK2054); Major Program from Xiangjiang Laboratory under Grant 23XJ02005; Central South University Research Programme of Advanced Interdisciplinary Studies (No.2023QYJC020).

%
%
%
\bibliographystyle{splncs04}
\bibliography{refs}

\end{document}